\documentclass[conference]{IEEEtran}
\IEEEoverridecommandlockouts

\usepackage[utf8]{inputenc}
\usepackage[T1]{fontenc}
\usepackage{cite} 
\usepackage{amsmath,amssymb,amsfonts}
\usepackage{graphicx}
\usepackage{textcomp}
\usepackage{xcolor}
\usepackage{booktabs}       
\usepackage{multirow}       
\usepackage{siunitx}        
\usepackage{subcaption}     
\usepackage{algorithm}      
\usepackage{algpseudocode}  
\usepackage{url}
\usepackage{tabularx}
\usepackage{ragged2e}
\usepackage{array}
\newcolumntype{C}{>{\centering\arraybackslash}X}
\usepackage{microtype}      

\usepackage{hyperref}
\hypersetup{
	colorlinks=true,
	linkcolor=black,
	citecolor=black,
	filecolor=black,
	urlcolor=black,
}

\def\BibTeX{{\rm B\kern-.05em{\sc i\kern-.025em b}\kern-.08em
		T\kern-.1667em\lower.7ex\hbox{E}\kern-.125emX}}

\begin{document}
	
	\title{Octopus Inspired Optimization (OIO): A Hierarchical Framework for Navigating Protein Fitness Landscapes}
	
	\author{\IEEEauthorblockN{
			Xu Wang\textsuperscript{\textit{a,d,*}},
			Yiquan Wang\textsuperscript{\textit{b,*}},
			Tin-Yeh Huang\textsuperscript{\textit{c}},
			Yuhua Dong\textsuperscript{\textit{a}},
			Jia Deng\textsuperscript{\textit{a}},
			Longji Xu\textsuperscript{\textit{a}},
			Xiang Li\textsuperscript{\textit{a}}, and
			Rui He\textsuperscript{\textit{a}}}
		\IEEEauthorblockA{\textsuperscript{\textit{a}}College of Communication Engineering, Jilin University, Changchun, China}
		\IEEEauthorblockA{\textsuperscript{\textit{b}}College of Mathematics and System Science, Xinjiang University, Urumqi, China}
		\IEEEauthorblockA{\textsuperscript{\textit{c}}Department of Industrial and Systems Engineering, The Hong Kong Polytechnic University, Hong Kong, China}
		\IEEEauthorblockA{\textsuperscript{\textit{d}}Tsinghua University-Peking University Joint Center for Life Sciences, Tsinghua University, Beijing, China}
		\IEEEauthorblockA{Email: wangxu2020@mails.jlu.edu.cn, ethan@stu.xju.edu.cn}
		\IEEEauthorblockA{\textsuperscript{*}These authors contributed equally to this work and are co-corresponding authors.}
	}
	
	\maketitle
	
	\begin{abstract}
		Navigating vast, rugged biological fitness landscapes to discover high-value functional patterns—such as optimal protein sequences—is a central challenge in health informatics. However, conventional algorithms often struggle with the exploration-exploitation dilemma, failing to synergize global search with deep local refinement, which leads to entrapment in suboptimal solutions. To overcome this barrier, we introduce Octopus Inspired Optimization (OIO), a novel hierarchical metaheuristic that mimics the octopus's unique neural architecture to intrinsically unify centralized global exploration and parallelized local exploitation. We validated OIO on a real-world protein engineering benchmark, where it surpassed 15 competing metaheuristics. This success is underpinned by OIO's architectural suitability for protein-like landscapes, confirmed by its top ranking on the NK-Landscape benchmark, and its powerful optimization engine, demonstrated by its first-place performance on the gold-standard CEC2022 benchmark. OIO thus provides a robust, nature-inspired computational tool for complex optimization problems in drug discovery and personalized medicine.
	\end{abstract}
	
	
	\section{Introduction}
	In the multi-omics era, the proliferation of technologies such as genomics and proteomics has revolutionized health informatics but also generated massive and complex high-dimensional data. A central challenge is to effectively extract meaningful biological insights from this data deluge, which is fundamentally a pattern mining (PM) problem. Recent approaches have demonstrated the efficacy of sequential pattern mining in this domain, successfully applying it to the classification of macromolecule genome sequences \cite{nawaz2024spm4gac}, the identification of frequent patterns in heat shock proteins \cite{nawaz2024fsp4hsp}, and the analysis of viral spike protein structures \cite{nawaz2022spdb}. Broadly, these challenges range from analytical tasks, like discovering patterns of genetic variation that lead to disease, to design tasks, such as engineering novel protein molecules with specific therapeutic functions. The essence of these tasks is a search and optimization problem, for which heuristic algorithms (HAs) are critical tools, set within a vast and rugged 'fitness landscape' \cite{romero2009exploring, romero2013navigating}—a high-dimensional space where each point represents a biological configuration and its altitude corresponds to its functional value. Protein engineering serves as a classic and critical example of this challenge in proteomics, a key domain within multi-omics, where the goal is to find an optimal amino acid sequence from a combinatorially astronomical search space \cite{ebrahimi2023engineering, ndochinwa2024new, albanese2025computational}. Navigating this landscape to find the global optimum is exceptionally difficult, and existing HAs often struggle with the "exploration-exploitation" dilemma \cite{sutton1998reinforcement, berger2014exploration}. Local search methods, like a mountaineer who can only see their immediate surroundings, are efficient at climbing the nearest hill (exploitation) but are easily trapped. Conversely, global search methods, such as classic HAs like Genetic Algorithms, act like reconnaissance aircraft, identifying many potential peaks (exploration) but lacking the tools to efficiently ascend each one \cite{hills2015exploration, taylor2022evolution, dokeroglu2019survey, jerebic2021novel}. What this demanding PM task truly requires is a system that synergistically combines reconnaissance and mountaineering—a strategy that can both globally survey the landscape for promising regions and locally mine those regions for the highest peaks in parallel.
	
	In recent years, deep learning methods, such as Protein Language Models (PLMs), have revolutionized protein engineering by serving as powerful surrogate fitness models or efficient sequence generators \cite{gelman2025biophysics, qiu2023artificial, brandes2022proteinbert, xu2020deep}. However, they are not optimization algorithms themselves; they leave a critical gap, as the fundamental challenge remains: how to efficiently guide the sequence search to find optimal solutions in the vast sequence space \cite{xu2020deep, nikolados2023deep, wu2021protein}. This is precisely where a more powerful search strategy is needed. Nature, through millennia of evolution, provides a perfect blueprint for such a strategy in the octopus. An octopus's brain formulates a global foraging plan, while its eight semi-autonomous arms act as independent, coordinated teams to explore, sense, and capture prey in parallel \cite{zullo2019motor, sivitilli2023mechanisms, hochner2006octopus}. This remarkable fusion of \textit{centralized global planning} and \textit{decentralized parallel execution} provides an ideal model for overcoming the exploration-exploitation dilemma inherent in protein fitness landscapes.
	
	Inspired by this biological blueprint, we introduce the Octopus Inspired Optimization (OIO), a novel metaheuristic algorithm that mimics this unique neural architecture. Our core contributions are twofold: 1) We propose a new algorithmic paradigm whose structure is naturally suited for solving complex combinatorial problems like protein sequence optimization. 2) We comprehensively validate OIO's performance through a three-tiered experimental framework: demonstrating its real-world efficacy on a protein design benchmark, confirming its structural fit with a classical landscape model, and verifying its robust underlying optimization engine on a standard benchmark suite. This work positions OIO as a powerful new tool not just for protein engineering, but for a broader range of optimization challenges in computational \textbf{drug discovery} and \textbf{personalized medicine}. Its advantages are demonstrated by systematically outperforming a comprehensive and diverse suite of algorithms, including both time-tested classic strategies and state-of-the-art metaheuristics from the last two years.
	
	\section{Related Works}
	\subsection{Computational Approaches for Protein Sequence Optimization}
	The computational search for optimal protein sequences is fundamentally constrained by the exploration-exploitation dilemma. Global strategies like conventional Evolutionary Algorithms (EAs) \cite{holland1992adaptation, sloss20202019, vikhar2016evolutionary, kong2023dynamic, kneiding2024augmenting} excel at surveying the vast fitness landscape but are often slow to converge, while Local Search and Heuristics like hill climbing rapidly ascend the nearest fitness peak but are easily trapped in suboptimal solutions \cite{selman2006hill, hong2024integrative, xue2025improving}. Each of these classes represents a point on the spectrum, forcing a trade-off. While hybrid approaches (e.g., Hybrid PSO-GWO \cite{shaikh2025intelligent}), often known as memetic algorithms, have sought to combine these strengths, they typically bolt on a local searcher to a global algorithm \cite{moscato1989evolution, neri2012memetic, hart2006recent, reddy2025advances, zhou2014compression}. A framework that intrinsically and structurally unifies high-level strategic exploration with deep, parallelized local exploitation remains a critical unmet need \cite{albanese2025computational, blanco2023role}. OIO is proposed not merely to integrate these mechanisms, but to embody this synergy within its core hierarchical architecture.
	
	\subsection{The Structural Limitations of General-Purpose Metaheuristics}
	To rigorously benchmark OIO, we compare it against a comprehensive suite of 15 metaheuristics. This suite was intentionally designed to be diverse, comprising 7 well-established, classic algorithms \cite{ selman2006hill, kirkpatrick1983optimization, kennedy1995particle, forrest1996genetic, storn1997differential, mirjalili2016whale, heidari2019harris} and 8 state-of-the-art optimizers \cite{shaikh2025intelligent, manzoor2023ahho, abdel2024crested, el2024greylag, al2024elk, amiri2024hippopotamus, al2024quokka, han2024walrus} proposed within the last two years. While this diverse set spans foundational paradigms from swarm intelligence (e.g., Particle Swarm Optimization, PSO \cite{kennedy1995particle}) to evolutionary methods (e.g., Differential Evolution, DE \cite{storn1997differential}), they largely share a critical structural limitation: a "flat" population structure. In this paradigm, all search agents operate at the same conceptual level, contributing to a shared global search  \cite{rajwar2023exhaustive, cai2024cooperative, dagal2025logarithmic}. They lack a native, hierarchical framework that explicitly decouples high-level strategic direction (like an octopus's brain) from parallel, semi-autonomous local refinement (like its tentacles). This structural flaw often forces the entire population to converge prematurely or maintain diversity at the cost of deep exploitation, highlighting the critical gap for an algorithm whose very architecture is designed for the multi-level search required by complex protein fitness landscapes \cite{osuna2022diversity, lim2013two, dokeroglu2024survey, turgut2023systematic}.
	
	\section{OIO: A Hierarchical Framework for Protein Optimization}
	
	\subsection{Biological Principles and the Hierarchical Synergy Framework}
\begin{figure}[htbp]
		\centering
		\includegraphics[width=0.8\columnwidth]{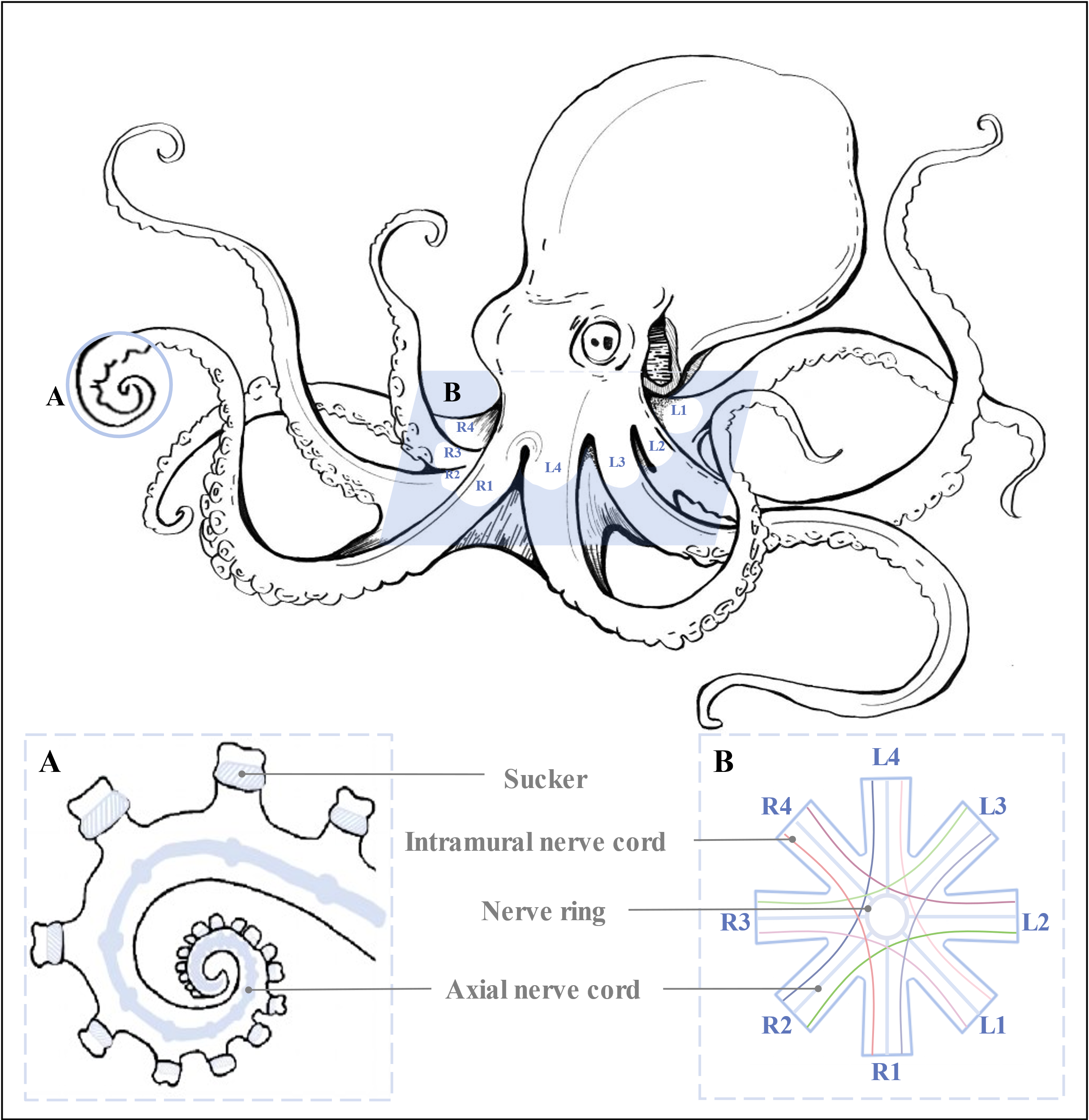}
		\caption{The octopus nervous system, illustrating the centralized brain and decentralized arm ganglia that inspire OIO's hierarchical architecture.}
		\label{fig:octopus_nervous_system}
	\end{figure}

	The design of OIO is a direct abstraction of the octopus's neurological and morphological features. We map this biological architecture to a hierarchical algorithmic framework, as shown in Fig.~\ref{fig:octopus_nervous_system}. This framework operates on three synergistic levels, each with a distinct role in navigating the protein fitness landscape.
	
\begin{figure*}[htb]
		\centering
		\includegraphics[width=\textwidth]{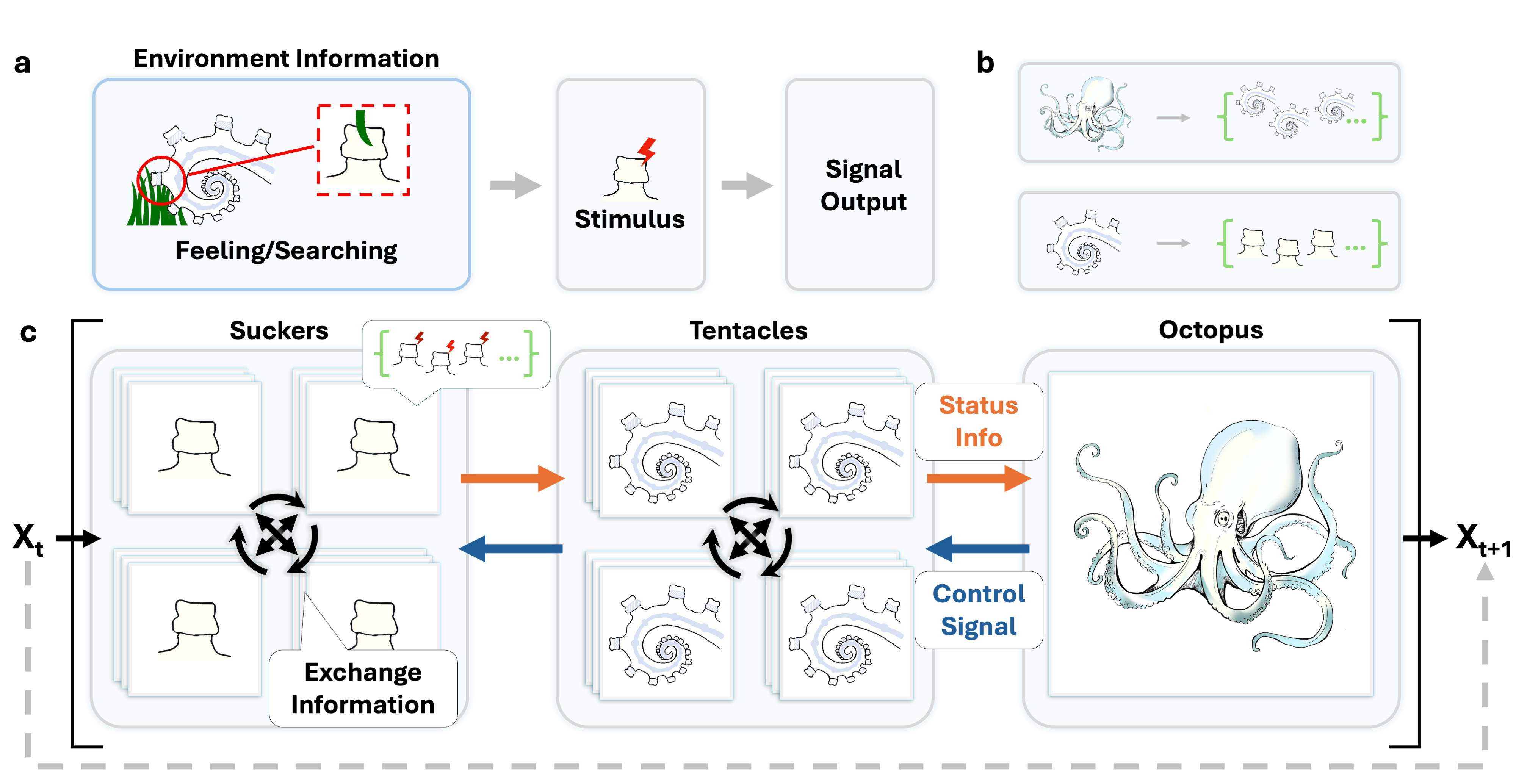}
		\caption{The simplified framework of OIO, integrating signal generation, hierarchical mapping, and information interaction. Panel (a) illustrates how suckers perceive the environment, generate stimuli, and output signals as the initial source of information. Panel (b) presents the biological hierarchy of the octopus abstracted into the algorithm, where the octopus corresponds to the global individual, tentacles act as regional agents, and suckers serve as local units. Panel (c) demonstrates the bidirectional flow of information across the three levels, where suckers exchange information and report status upward, tentacles aggregate and transmit results to the octopus, and the octopus integrates feedback and issues control signals downward to guide the next iteration.}
		\label{fig:oio_hierarchy}
	\end{figure*}

	This hierarchical structure, inspired by the octopus's nervous system, is visualized in Fig.~\ref{fig:oio_hierarchy}. The framework's power lies in its clear mapping of biological roles to algorithmic components and, most critically, in the dynamic, bidirectional flow of information that drives the optimization. This feedback loop—where local data is aggregated upward and global strategy is disseminated downward—creates a synergistic system that intrinsically balances exploration and exploitation.
	
	\subsection{Core Optimization Mechanisms}
\begin{figure*}[htb]
		\centering
		\includegraphics[width=\textwidth]{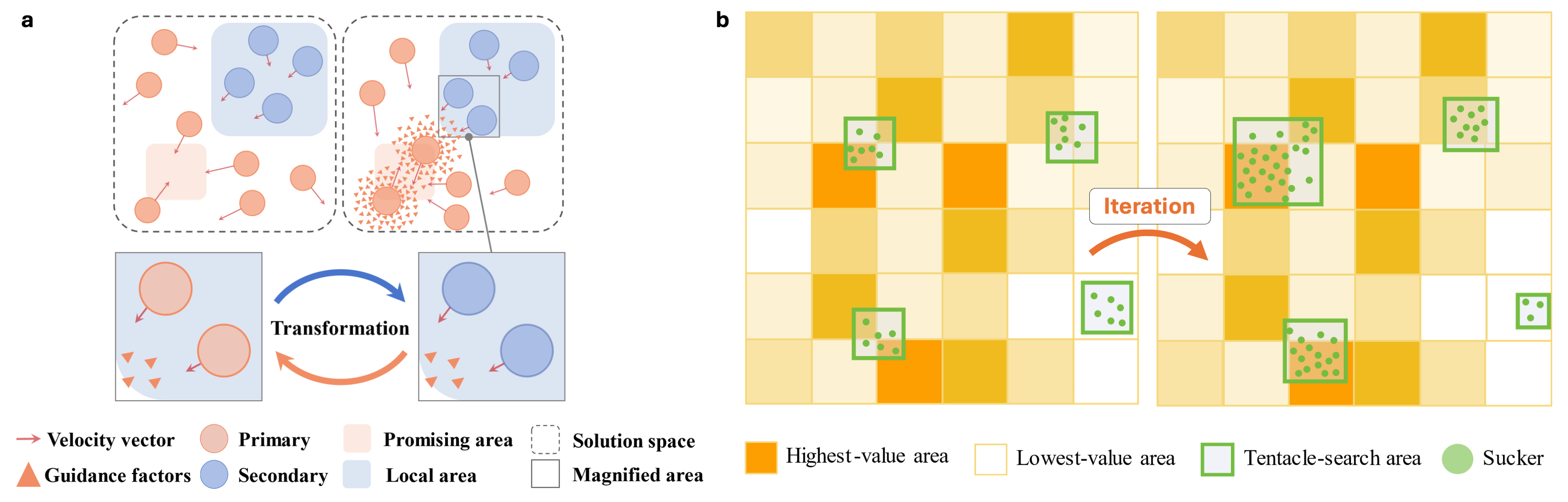}
		\caption{Core dynamic mechanisms of OIO. Panel (a) illustrates the Group Co-evolution strategy, which balances exploration and exploitation using master and slave tentacles and includes an adaptive regeneration mechanism to escape local optima. Panel (b) shows the Iterative Optimization Process, a dynamic feedback loop where information flows from the suckers up to the individual, enabling adaptive resource allocation.}
		\label{fig:oio_dynamics}
	\end{figure*}

	Two core mechanisms, illustrated in Fig.~\ref{fig:oio_dynamics}, drive OIO's search process and ensure a robust balance between exploration and exploitation.
	
	\textbf{Co-evolution and Adaptive Regeneration.} OIO employs a group co-evolution strategy where some tentacles act as 'masters' to deeply exploit known high-fitness regions, while others act as 'slaves' to explore new areas (Fig.~\ref{fig:oio_dynamics}(a)). Critically, an underperforming tentacle can be adaptively regenerated to a new random location. This is a key mechanism for escaping the numerous local optima characteristic of protein fitness landscapes.
	
	\textbf{The Iterative Optimization Process.} The entire optimization unfolds as a dynamic, iterative process (Fig.~\ref{fig:oio_dynamics}(b)). In each cycle, a multi-level feedback loop occurs: suckers perform local mutations, tentacles aggregate this information to adjust their regional search strategy, and the individual oversees the global picture, potentially reassigning tentacles. This process allows OIO to dynamically allocate computational resources to the most promising areas of the fitness landscape.
	
	\subsection{Mathematical Formulation of Core Mechanisms}
	To formally define the optimization dynamics within the OIO framework, we present the core mathematical equations governing the behavior of the Suckers and Tentacles.
	
	At the heart of OIO's local exploitation is the position update for each Sucker. Let $\vec{S}_{ij}(t)$ be the position of the $j$-th Sucker of the $i$-th Tentacle at iteration $t$. Its new position $\vec{S}_{ij}(t+1)$ is determined by a hybrid strategy that adaptively switches between exploration and exploitation. This transition is governed by a dynamic energy factor $E$. At iteration $t$ (up to a maximum $T_{\text{max}}$), $E$ is calculated using a random seed $E_0 \in [-1, 1]$:
	\begin{equation}
		E = 2 E_0 (1 - t/T_{\text{max}})
		\label{eq:energy}
	\end{equation}
	The primary update rule depends on the magnitude of $E$:
	\begin{equation}
		\vec{S}_{ij}(t+1) = 
		\begin{cases} 
			\begin{aligned} 
				& \text{Exploration Strategy} \\
				& (\vec{S}_{ij}(t), \vec{S}_{\text{rand}}, \dots)
			\end{aligned} 
			& \text{if } |E| \geq 1 \\

			\begin{aligned} 
				& \text{Exploitation Strategy} \\
				& (\vec{S}_{ij}(t), \vec{T}_{i,\text{best}}, \vec{G}_{\text{best}})
			\end{aligned} 
			& \text{if } |E| < 1 
		\end{cases}
		\label{eq1} 
	\end{equation}
	where $\vec{T}_{i,\text{best}}$ is the best-known position within Tentacle $T_i$, and $\vec{G}_{\text{best}}$ is the global best position found by the entire individual. The exploitation strategy synthesizes information from the Sucker's personal best, the Tentacle's best, and the global best, resembling a multi-level social learning structure inspired by Particle Swarm Optimization (PSO):
	\begin{align}
		\vec{v}_{ij}(t+1) = {} & w\vec{v}_{ij}(t) + c_1 r_1 (\vec{P}_{ij,\text{best}} - \vec{S}_{ij}(t)) \nonumber \\ 
		& + c_2 r_2 (\vec{T}_{i,\text{best}} - \vec{S}_{ij}(t)) \nonumber \\
		& + c_3 r_3 (\vec{G}_{\text{best}} - \vec{S}_{ij}(t))
		\label{eq2}
	\end{align}
	\begin{equation}
		\vec{S}_{ij}(t+1) = \vec{S}_{ij}(t) + \vec{v}_{ij}(t+1)
		\label{eq3}
	\end{equation}
	Here, $\vec{v}_{ij}$ is the velocity vector, $w, c_1, c_2, c_3$ are inertial and acceleration coefficients, and $r_1, r_2, r_3$ are random numbers. This formulation extends canonical PSO by incorporating the Tentacle's collective intelligence ($\vec{T}_{i,\text{best}}$), creating a more structured and efficient local search.
	
	The adaptive regeneration mechanism is triggered for a Tentacle $T_i$ when its stagnation counter, $\text{Stag}_i$, exceeds a predefined threshold $\tau_{\text{stag}}$. Regeneration involves repositioning the Tentacle's search center $\vec{C}_i$:
	\begin{equation}
		\text{if } \text{Stag}_i > \tau_{\text{stag}} \quad \text{then} \quad \vec{C}_i(t+1) = \vec{L} + \text{rand}() \cdot (\vec{U} - \vec{L})
		\label{eq4}
	\end{equation}
	where $\vec{L}$ and $\vec{U}$ define the lower and upper bounds of the search space. The complete procedure is outlined in Algorithm~\ref{alg:oio}.
	
	\begin{algorithm}[t]
		\caption{The Octopus Inspired Optimization (OIO)}
		\label{alg:oio}
		\begin{algorithmic}[1]
			\State \textbf{Input:} Fitness function $f(x)$, Max Iterations $T_{\text{max}}$
			\State \textbf{Initialize:} Individual $I$, Tentacles $\{T_1, ..., T_m\}$, Suckers $\{S_{ij}\}$
			\State Evaluate all Suckers' initial fitness
			\While{$t < T_{\text{max}}$}
			\State Calculate Energy Factor $E$ via Eq.~(\ref{eq:energy})
			\ForAll{Tentacle $T_i$ in $\{T_1, ..., T_m\}$}
			\ForAll{Sucker $S_{ij}$ in $T_i$}
			\State Update position via Eq.~(\ref{eq1})-(\ref{eq3}) based on $E$
			\State Evaluate fitness and update personal best $\vec{P}_{ij,\text{best}}$
			\EndFor
			\State Update Tentacle's best $\vec{T}_{i,\text{best}}$
			\EndFor
			\State Update global best solution $\vec{G}_{\text{best}}$
			\State Perform regeneration (Eq.~\ref{eq4}) if $\text{Stag}_i > \tau_{\text{stag}}$
			\State $t \leftarrow t + 1$
			\EndWhile
			\State \textbf{Return:} The best solution $\vec{G}_{\text{best}}$
		\end{algorithmic}
	\end{algorithm}

\subsection{Encoding Strategy for Protein Sequence Optimization}
A critical component for applying OIO to protein engineering is the strategy for encoding a discrete protein sequence within the algorithm's continuous internal search space. We adopt a probabilistic matrix encoding scheme. For a protein of length $L$, an internal solution in OIO is represented by a one-dimensional vector of length $L \times 20$. During fitness evaluation, this vector is reshaped into an $L \times 20$ matrix $\mathbf{P}$, where the element $P_{ij}$ represents a preference score for the $j$-th amino acid (out of 20) appearing at the $i$-th position in the sequence.

Subsequently, a Softmax function is applied to each row of this matrix, converting the scores into a valid probability distribution. The final protein sequence is then generated by sampling an amino acid for each position based on these probabilities (or by deterministically selecting the amino acid with the highest probability). This encoding allows OIO to perform smooth optimization in a continuous internal space while effectively exploring the discrete space of protein sequences.

\section{Experiments and Analysis}
\subsection{Experimental Design}
To comprehensively validate the OIO algorithm, we designed a rigorous three-tiered experimental framework intended to dissect its performance from practical application down to its core mechanics. This framework is guided by three progressively deeper questions. First, we present a case study of HA application in proteomics by assessing OIO's real-world efficacy on the demanding DMS-GFP protein optimization task \cite{sarkisyan2016local}. Second, we investigate its structural affinity for protein-like fitness landscapes using the classic NK-Landscape model to understand the architectural reasons for its success \cite{kauffman1992origins}. Finally, we evaluate its core engine performance using the gold-standard CEC2022 benchmark suite to confirm the fundamental power and robustness of its optimization mechanisms \cite{Kumar2021CEC2022, CEC2022Competition}.

Across all three benchmarks, OIO was compared against a comprehensive and challenging suite of 15 metaheuristic algorithms, comprising both classic methods and state-of-the-art optimizers. To ensure reproducibility, the specific parameter settings for OIO are detailed in Table~\ref{tab:oio_params}. This rigorous approach ensures that OIO's performance is validated against a diverse and contemporary set of optimization strategies, confirming its novelty and effectiveness.

\begin{table}[htbp]
	\centering
	\caption{Hyperparameter settings for the OIO algorithm.}
	\label{tab:oio_params}
	\begin{tabular}{lc}
		\toprule
		\textbf{Hyperparameter} & \textbf{Value} \\
		\midrule
		Number of Tentacles & 5 \\
		Suckers per Tentacle & 40 \\
		Iterations per Tentacle & 100 \\
		Stagnation Threshold for Regeneration & 5 iterations \\
		Diversity Threshold & 0.005 \\
		Elite Memory Size & 8 \\
		\bottomrule
	\end{tabular}
\end{table}

\subsection{Core Validation: Real-World GFP Optimization}
\begin{figure}[t]
\centering
\includegraphics[width=\columnwidth]{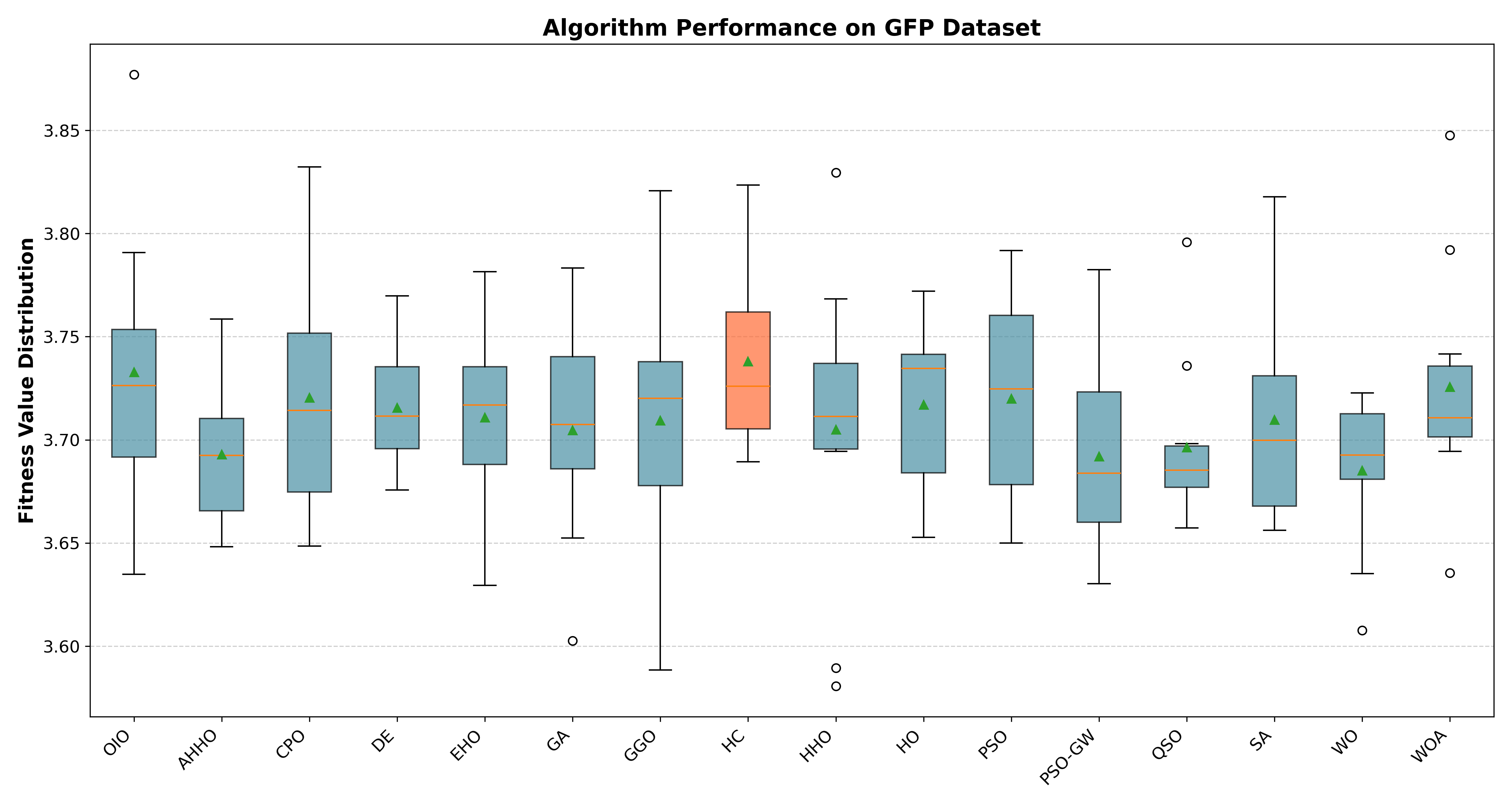}
\caption{Performance comparison of OIO and 15 baseline algorithms on the GFP optimization task. The boxplot illustrates the distribution of final fitness values across 10 independent runs. OIO (highlighted) demonstrates a superior median fitness and a consistently high-performing distribution, second only to the specialist Hill Climbing algorithm.}
\label{fig:dms_boxplot}
\end{figure}

The primary validation of OIO was conducted on the GFP sequence optimization task, where the goal is to find the amino acid sequence with the highest fluorescence intensity. The results, visualized in Fig.~\ref{fig:dms_boxplot}, are highly illuminating.

The most striking result is that OIO achieved the second-best overall rank among 16 algorithms, surpassed only by Hill Climbing (HC). Crucially, OIO was the top-performing algorithm among all general-purpose metaheuristics. This outcome is significant and requires careful interpretation. Hill Climbing is a pure local search method—a specialized "exploiter." Its success confirms that the GFP fitness landscape contains steep, climbable peaks. However, HC's performance is entirely dependent on its starting point; from a poor initial sequence, it would be irreversibly trapped in a suboptimal region.

OIO, by contrast, is a general-purpose optimizer designed to balance exploration and exploitation. The fact that it achieved a performance nearly identical to a specialized exploiter, without sacrificing its global search capability, is a powerful testament to the efficiency of its "tentacle-sucker" mechanism for deep, parallel local refinement. This makes OIO far more robust for unknown and potentially more complex fitness landscapes where a good starting point is not guaranteed. As illustrated by the tighter high-fitness distribution in Fig.~\ref{fig:dms_boxplot}, OIO consistently achieved a higher mean fitness than 14 out of the 15 other algorithms, demonstrating a superior balance of global robustness and local efficiency—a critical requirement for practical protein engineering tasks. Furthermore, the search trajectory indicates OIO rapidly approaches the high-fitness region found by the specialist HC algorithm. This complementary nature suggests a promising future research direction: employing OIO for robust global exploration to identify high-potential basins of attraction, followed by a rapid HC-like local search for final, efficient refinement. This hybrid strategy could potentially achieve the best of both worlds.

\subsection{Structural Affinity Analysis: The NK-Landscape Benchmark}
The NK-Landscape is a tunable mathematical model designed specifically to replicate the "rugged" and correlated nature of protein fitness landscapes. An algorithm that performs well on this benchmark is likely to have a search strategy that is intrinsically compatible with this type of problem structure. To ensure a rigorous evaluation, we constructed a test suite of five NK configurations with increasing difficulty, from Simple (N=20, K=2) to Complex (N=100, K=5). For each configuration, every algorithm was run 10 times with a budget of 20,000 function evaluations, and all continuous-space optimizers used a unified Sigmoid transfer function for a fair comparison.

\begin{figure}[t]
\centering
\includegraphics[width=\columnwidth]{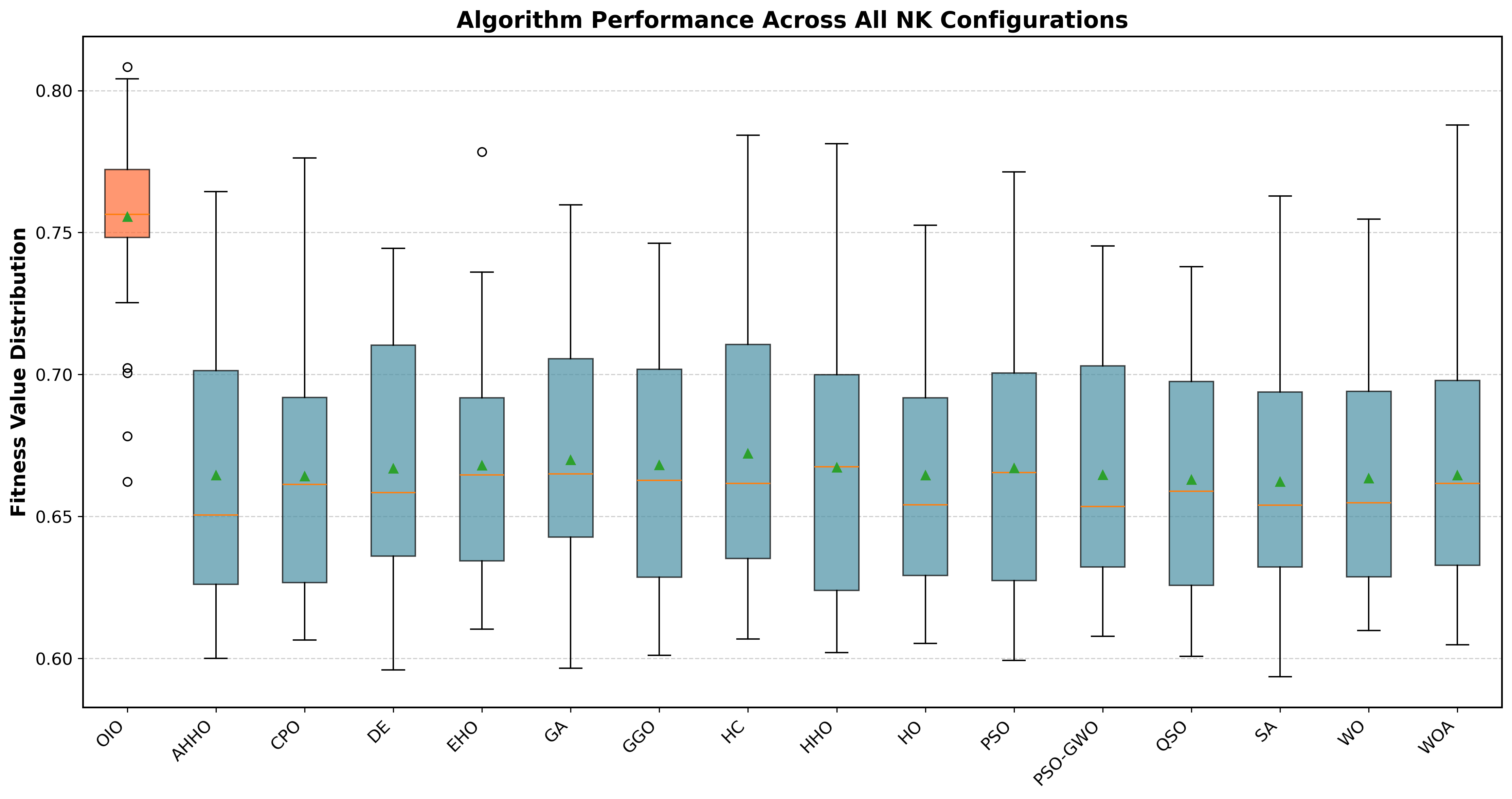}
\caption{Comprehensive performance comparison on the NK-Landscape benchmark. The boxplot shows the distribution of final fitness values from 10 independent runs across all five NK configurations. OIO (highlighted) exhibits a markedly superior distribution, with a higher median and less variance compared to the 15 baseline algorithms.}
\label{fig:nk_boxplot}
\end{figure}

The results, summarized in Fig.~\ref{fig:nk_boxplot}, are decisive. OIO achieved the number one overall rank among all 16 algorithms. The boxplot in Fig.~\ref{fig:nk_boxplot} visually confirms this dominance, showing that OIO's performance distribution is clearly superior to all other competitors, with a consistently higher median and lower variance. Furthermore, quantitative analysis reveals that OIO consistently obtained a higher mean fitness than the entire cohort of baseline algorithms across all five levels of landscape complexity. This commanding performance provides strong evidence that OIO's hierarchical, parallel search architecture is naturally suited to navigating the high-dimensional, multi-modal, and rugged landscapes characteristic of protein optimization, explaining its ability to effectively map and search the complex GFP fitness landscape. This structural advantage translates into superior search efficiency, where OIO not only finds a better final solution but also converges significantly faster than the other top-performing algorithms.

\subsection{Core Engine Validation: The CEC2022 Benchmark}
While OIO operates on discrete protein sequences, its internal search process, guided by the probabilistic encoding matrix, occurs in a continuous space. The CEC2022 benchmark suite is the standard for evaluating the core performance of continuous optimizers, rigorously testing their ability to balance global exploration and local exploitation. For this validation, all 16 algorithms were tested on the 12 standard 10-dimensional problems (F1-F12). Each was run 10 times with a maximum budget of 20,000 function evaluations (FES).

The results unequivocally demonstrate the superiority of OIO's core engine. Quantitative ranking analysis confirms that OIO achieved the number one overall rank, significantly outperforming the second-place CPO. This commanding lead confirms OIO's world-class optimization power.

\begin{figure}[t]
\centering
\includegraphics[width=\columnwidth]{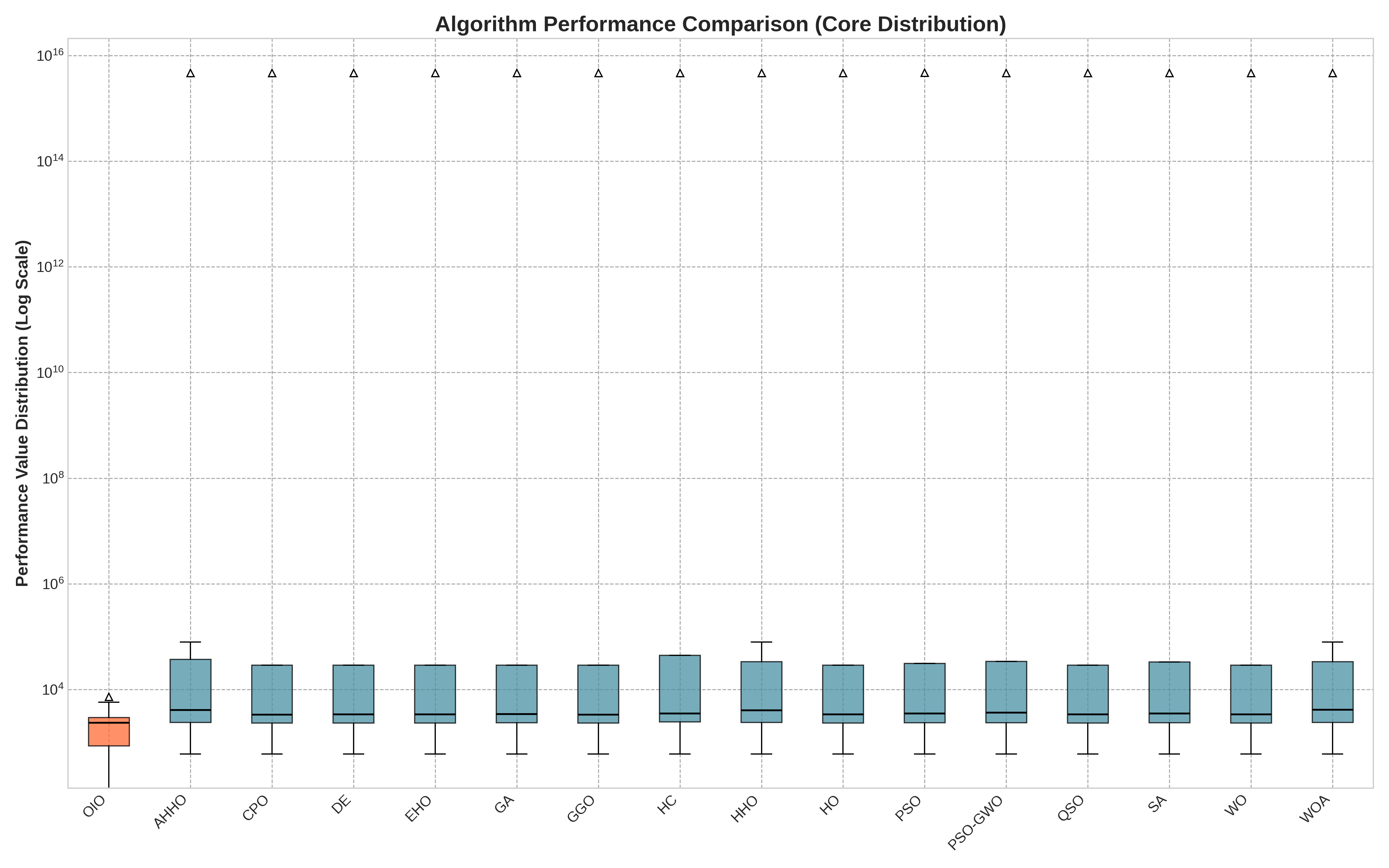}
\caption{Performance comparison on the CEC2022 benchmark suite. The boxplot shows the distribution of final fitness values from 10 independent runs across all 12 functions. OIO (highlighted) demonstrates a substantially lower median and a more compact distribution, indicating superior performance and higher stability.}
\label{fig:cec_boxplot}
\end{figure}

The performance gap is visually evident in Fig.~\ref{fig:cec_boxplot}. OIO's results (highlighted) not only exhibit a lower median, indicating better solutions, but also a much tighter interquartile range, signifying exceptional stability. Table~\ref{tab:cec_summary} further quantifies this, showing that on multimodal problems like F6, OIO finds solutions orders of magnitude better than the best baseline, showcasing its critical ability to escape local optima. In terms of convergence behavior, OIO exhibits a rapid descent towards superior solutions across unimodal, multimodal, and composition functions, starkly contrasting with the slower convergence or premature stagnation observed in other leading algorithms.

\begin{table}[htbp]
\centering
\caption{Mean Fitness Comparison on Representative CEC2022 Functions.}
\label{tab:cec_summary}
\sisetup{scientific-notation = true, round-mode = places, round-precision = 2}
\begin{tabular}{lccc}
	\toprule
	\textbf{Function} & \textbf{OIO Mean} & \textbf{Best Baseline} & \textbf{Avg. of} \\
	& \textbf{Fitness} & \textbf{Mean} & \textbf{Baselines} \\
	\midrule
	F1 (Unimodal) & \textbf{\num{4.10e3}} & \num{7.99e4} (CPO) & \num{1.12e5} \\
	F6 (Multimodal) & \textbf{\num{6.57e4}} & \num{2.31e9} (CPO) & \num{2.90e9} \\
	F12 (Composition) & \textbf{\num{2.93e3}} & \num{2.91e3} (CPO) & \num{2.95e3} \\
	\bottomrule
\end{tabular}
\end{table}

Beyond solution quality, OIO excels in computational efficiency. As shown in Table~\ref{tab:cec_time_summary}, OIO consistently finds its superior solutions in a significantly shorter timeframe. On every representative function, OIO's average execution time is substantially lower than even the fastest baseline algorithm. For example, on the complex F6 function, OIO is over 30\% faster than the quickest competitor (34.3s vs 48.8s) and nearly twice as fast as the baseline average. This demonstrates that OIO's performance gains do not come at the cost of computational overhead; rather, its intelligent search strategy leads to a more efficient discovery of high-quality solutions. This combination of accuracy, stability, and speed solidifies OIO's position as a powerful and practical optimization tool.

\begin{table}[htbp]
\centering
\caption{Comparison of Mean Execution Time (seconds) on Representative CEC2022 Functions.}
\label{tab:cec_time_summary}
\begin{tabular}{lcccc}
	\toprule
	\textbf{Function} & \textbf{OIO} & \textbf{Fastest} & \textbf{Avg.} & \textbf{Slowest} \\
	& \textbf{Avg. Time} & \textbf{Baseline} & \textbf{Baseline} & \textbf{Baseline} \\
	\midrule
	F1 & \textbf{22.49s} & 32.33s & 43.60s & 61.25s \\
	F6 & \textbf{34.26s} & 48.82s & 63.87s & 86.85s \\
	F12 & \textbf{148.77s} & 207.46s & 257.57s & 332.03s \\
	\bottomrule
\end{tabular}
\end{table}

\section{Discussion \& Conclusion}
In this work, we addressed the significant challenge of computational protein engineering by proposing the Octopus Optimizer (OIO), a novel metaheuristic algorithm inspired by the sophisticated neural system of the octopus. Through a three-tiered experimental framework, we have demonstrated that: 1) OIO exhibits top-tier practical performance on a real-world GFP protein design task; 2) its success is rooted in a fundamental architectural alignment with the nature of protein fitness landscapes, as validated on the NK-Landscape benchmark; and 3) its performance is powered by a state-of-the-art optimization engine, confirmed by its number one ranking on the CEC2022 benchmark suite.

The core advantage of OIO lies in its intrinsic, structured synergy between exploration and exploitation. This makes it more robust and efficient than traditional methods for complex black-box problems, such as protein optimization, which demand both a global perspective and deep local refinement. Furthermore, its architecture with multiple semi-autonomous tentacles is naturally suited for distributed and parallel HA frameworks, addressing the critical need for scalability when tackling large-scale multi-omics data.

The significance of this work, however, extends beyond outperforming other algorithms; it lies in establishing OIO's strategic role within modern health informatics. We identify two core ecological niches for OIO. First, it can form powerful hybrid approaches combining HA with Machine Learning (ML), a key theme of this workshop. In such a synergy, OIO acts as an intelligent 'Optimization Driver' proposing candidates for a DL-based 'Evaluator' (e.g., a PLM), combining OIO's search power with the domain knowledge of DL models. Second, in many 'black-box' scenarios central to health informatics—such as early-stage drug discovery where wet-lab experiments are costly, or designing therapies for rare diseases where training data is scarce—DL models are not viable \cite{ching2018opportunities, sapoval2022current, van2024deep}. In these critical situations, an efficient, gradient-free optimizer like OIO is indispensable \cite{audet2016blackbox}.

This positioning opens up new avenues for computational biology. Future work will focus on four key directions: 1) extending OIO to handle multi-objective protein design \cite{hong2024integrative, tokuriki2008protein, mistani2024preference, stern2023probabilistic}; 2) applying it to \textit{de novo} drug discovery and molecular design \cite{schneider2018automating, jumper2021highly, zeng2022deep}; 3) adapting the OIO framework to other multi-omics optimization problems, such as identifying optimal biomarker combinations from transcriptomics data for clinical decision support, or optimizing microbial compositions in microbiome analysis; and 4) pursuing the deep integration of OIO with PLMs to create practical, high-performance optimization systems. Ultimately, OIO provides a powerful new tool, demonstrating that nature-inspired computational strategies are vital for solving humanity's most complex scientific challenges.

\section*{Acknowledgment}
The source code for this paper is available on GitHub: \url{https://github.com/JLU-WangXu/Octopus_Inspired_Optimization_OIO}.

\bibliographystyle{IEEEtran}
\bibliography{references}

\end{document}